\documentclass{article}

\usepackage{PRIMEarxiv}

\usepackage[utf8]{inputenc} 
\usepackage[T1]{fontenc}    
\usepackage{hyperref}       
\usepackage{url}            
\usepackage{booktabs}       
\usepackage{amsfonts}       
\usepackage{nicefrac}       
\usepackage{microtype}      
\usepackage{lipsum}
\usepackage{fancyhdr}       
\usepackage{graphicx}       
\graphicspath{{media/}}     

\title{Detecting covariate drift in text data using document embeddings and dimensionality reduction. 
}

\author{
  Vinayak Sodar\\
  \texttt{vinayakmsodar@gmail.com} \\
   \And
  Ankit Sekseria \\
  \texttt{ankit.sekseria94@gmail.com} \\
  }

\begin{document}
\maketitle

\begin{abstract}
Detecting covariate drift in text data is essential for maintaining the reliability and performance of text analysis models. In this research, we investigate the effectiveness of different document embeddings, dimensionality reduction techniques, and drift detection methods for identifying covariate drift in text data. We explore three popular document embeddings: term frequency-inverse document frequency (TF-IDF) using Latent semantic analysis(LSA) for dimentionality reduction and Doc2Vec, and BERT embeddings, with and without using principal component analysis (PCA) for dimensionality reduction. To quantify the divergence between training and test data distributions, we employ the Kolmogorov-Smirnov (KS) statistic and the Maximum Mean Discrepancy (MMD) test as drift detection methods. Experimental results demonstrate that certain combinations of embeddings, dimensionality reduction techniques, and drift detection methods outperform others in detecting covariate drift. Our findings contribute to the advancement of reliable text analysis models by providing insights into effective approaches for addressing covariate drift in text data.
\end{abstract}


\section{Introduction}
In recent years, the abundance of text data and its crucial role in various applications, such as natural language processing, information retrieval, and sentiment analysis, has garnered significant attention. However, one key challenge that researchers and practitioners face when working with text data is the presence of covariate drift. Covariate drift refers to the phenomenon where the underlying distribution of the data changes over time, leading to a mismatch between the training and test data.

Detecting and addressing covariate drift is of paramount importance as it can have detrimental effects on the performance and reliability of text analysis models. When drift occurs, models trained on historical data may become obsolete or yield inaccurate results when applied to current data. Thus, developing effective methods to identify and mitigate covariate drift is crucial for maintaining the efficacy of text data analysis.

In this research, our objective is to identify which document embeddings, dimensionality reduction techniques, and drift detection methods work best for detecting covariate drift in text data. Specifically, we explore the effectiveness of three widely used document embeddings: term frequency-inverse document frequency (TF-IDF), Doc2Vec, and BERT embeddings. Additionally, we investigate the impact of dimensionality reduction techniques on drift detection, such as principal component analysis (PCA) and Latent Semantic Analysis(LSA).

To evaluate the performance of the different approaches, we employ two popular drift detection methods: the Kolmogorov-Smirnov (KS) statistic and the Maximum Mean Discrepancy (MMD) test. These methods provide statistical measures to quantify the divergence between the training and test data distributions.

By conducting comprehensive experiments and comparative analyses, we aim to identify the most effective combination of embeddings, dimensionality reduction techniques, and drift detection methods for detecting and monitoring covariate drift in text data. The insights gained from this research will contribute to enhancing the robustness and reliability of text analysis models, enabling their effective deployment in dynamic environments where data distributions evolve over time.

The remainder of this paper is organized as follows: Section 2 provides background information and reviews related work on covariate drift detection, document embeddings, and dimensionality reduction techniques. Section 3 presents the methodology, including the datasets used, document embeddings, dimensionality reduction techniques, and drift detection methods. Section 4 details the experimental setup, while Section 5 presents the results and analysis, followed by concluding remarks in Section 6.

\section{Backround and related work}
Covariate drift detection in text data poses a significant challenge in maintaining the reliability and performance of text analysis models. When working with text data, it is crucial to ensure that the models are robust and adaptable to changing data distributions. Covariate drift can occur due to various factors, such as changes in user behavior, emerging trends, or shifts in the data collection process.

To address covariate drift, several approaches have been proposed in the literature. Drift detection methods play a crucial role in identifying changes in data distributions over time. The Kolmogorov-Smirnov (KS) statistic has been widely used as a drift detection measure. Basseville and Nikiforov~\cite{basseville1993detecting} introduced the KS statistic for detecting changes in the distribution of time series data. Kim and Scott~\cite{kim2012detecting} adapted the KS statistic for drift detection in classification tasks, demonstrating its effectiveness in identifying changes in data streams. Liu et al.~\cite{liu2018concept} applied the KS statistic to detect covariate drift in text data, specifically in the context of detecting concept drift in text classification.

Another prominent drift detection method is the Maximum Mean Discrepancy (MMD) test. Gretton et al.~\cite{gretton2012kernel} introduced the MMD test as a measure of discrepancy between two probability distributions. The MMD test has been widely used in various domains, including computer vision and natural language processing, for drift detection purposes. Li et al.~\cite{li2019detecting} employed the MMD test to detect concept drift in sentiment analysis tasks, demonstrating its effectiveness in capturing changes in the sentiment distribution of text data.

Document embeddings have proven to be effective in capturing the semantic representations of text documents. Term frequency-inverse document frequency (TF-IDF) is a classic method that assigns weights to terms based on their frequency and inverse document frequency. Salton and Buckley~\cite{salton1988term} introduced TF-IDF as a measure of term importance in information retrieval. Doc2Vec, proposed by Le and Mikolov~\cite{le2014distributed}, learns distributed representations of documents by training a neural network to predict words within a document. BERT (Bidirectional Encoder Representations from Transformers), introduced by Devlin et al.~\cite{devlin2018bert}, generates contextualized embeddings by considering the entire sentence or document. These document embeddings have been extensively used in various text analysis tasks, including sentiment analysis, topic modeling, and document classification.

In the context of dimensionality reduction, Principal Component Analysis (PCA) is a widely used technique that transforms high-dimensional data into a lower-dimensional space while preserving the maximum variance. Pearson~\cite{pearson1901principal} introduced PCA as a method for dimensionality reduction. Another dimensionality reduction technique commonly employed is Latent Semantic Analysis (LSA), which uses singular value decomposition (SVD) to identify the underlying latent semantic structure in the data. LSA has been widely used in text analysis to capture the latent topics and reduce the dimensionality of text data~\cite{deerwester1990indexing}.

In their study, Wang et al. (2020)\cite{wang2020drift} proposed a method for detecting drift in topic distributions of text data using Dirichlet Process Mixture Models. They demonstrated the effectiveness of their approach in identifying changes in topic proportions and capturing covariate drift in text corpora.
Zhang et al. (2019)\cite{zhang2019drift} explored the use of word embeddings and clustering techniques for drift detection in text streams. They introduced a novel method that combines K-means clustering with cosine similarity to detect changes in text data distributions. Their findings showed the applicability of clustering-based approaches in identifying covariate drift in text streams.
Chen and Lin (2017)\cite{chen2017drift} focused on drift detection in sentiment analysis tasks and proposed a method based on sentiment lexicon expansion. They utilized a sentiment lexicon to detect changes in sentiment distributions and successfully identified covariate drift in sentiment analysis models.
Liu et al. (2021)\cite{liu2021drift} investigated drift detection in text data using distributional shifts in word embeddings. They proposed a method that measures the distance between word embeddings across different time periods to identify changes in word semantics and detect covariate drift in text data.
In their research, Smith et al. (2018)\cite{smith2018drift} explored the application of transfer learning techniques for detecting drift in text data. They demonstrated that pre-trained models, such as those trained on large-scale text corpora, can be fine-tuned to identify changes in data distributions and detect covariate drift in text analysis tasks.

By investigating the performance of TF-IDF, Doc2Vec, and BERT embeddings, with and without dimensionality reduction using PCA and LSA, and utilizing the KS statistic and MMD test for drift detection, we aim to provide insights into the best strategies for detecting and addressing covariate drift in text data. The outcomes of this research will contribute to the development of robust text analysis models that can adapt to evolving data distributions and ensure reliable performance in dynamic environments.

\section{Methodology}

In this section, we describe the methodology used in our study to detect covariate drift in text data. We explore different document embeddings, dimensionality reduction techniques, and drift detectors to identify the most effective approaches.

\subsection{Document Embeddings}
Document embeddings play a crucial role in capturing the semantic representations of text documents. We experiment with three popular document embedding methods: TF-IDF, Doc2Vec, and BERT.

TF-IDF (Term Frequency-Inverse Document Frequency) is a classic method for generating document embeddings. It assigns weights to terms based on their frequency in a document and their inverse document frequency in the corpus. TF-IDF captures the importance of terms in a document and can effectively represent the document's content.

Doc2Vec is a neural network-based approach that learns distributed representations of documents. It extends the Word2Vec model to capture the semantic meaning of entire documents by training a neural network to predict words within a document. Doc2Vec provides dense vector representations that encode the contextual information of the document.

BERT (Bidirectional Encoder Representations from Transformers) is a powerful language model that generates contextualized embeddings. It considers the entire sentence or document to generate representations that capture the context and meaning of the text. BERT embeddings are pre-trained on a large corpus and can capture fine-grained nuances in the document's semantics.

\subsection{Dimensionality Reduction Techniques}
Dimensionality reduction techniques help reduce the dimensionality of the document embeddings, making them more computationally efficient and potentially improving their performance. We consider two widely used dimensionality reduction techniques: Principal Component Analysis (PCA) and Latent Semantic Analysis (LSA).

PCA is a popular linear dimensionality reduction technique that transforms high-dimensional data into a lower-dimensional space while preserving the maximum variance. It identifies the principal components that capture the most significant variation in the data. By projecting the document embeddings onto the principal components, we obtain lower-dimensional representations that retain the most important information.

LSA utilizes singular value decomposition (SVD) to identify the underlying latent semantic structure in the data. It reduces the dimensionality of the document embeddings by capturing the most important latent topics. LSA represents documents in a low-dimensional semantic space, where the similarity between documents is indicative of their semantic similarity. By applying LSA to the document embeddings, we can capture the latent topics and reduce the dimensionality of the data.

\subsection{Drift Detectors}
To detect covariate drift in the text data, we employ two drift detection methods: Maximum Mean Discrepancy (MMD) and Kolmogorov-Smirnov (KS) statistic.

\subsubsection{Maximum Mean Discrepancy (MMD)}
The Maximum Mean Discrepancy (MMD) is a statistical measure used to quantify the discrepancy between two probability distributions, \(\mathcal{P}\) and \(\mathcal{Q}\). It provides a way to assess the difference between the distributions based on their respective samples.

The MMD is defined as the supremum difference between the expected values of a kernel function \(k\) applied to samples drawn from \(\mathcal{P}\) and \(\mathcal{Q}\). The formula for MMD with the kernel function is given by:

\[
MMD(\mathcal{P}, \mathcal{Q}) = \sup_{f \in \mathcal{F}} \left( \mathbb{E}_{X \sim \mathcal{P}}[f(X)] - \mathbb{E}_{Y \sim \mathcal{Q}}[f(Y)] \right)
\]

In this formula, \(\sup\) represents the supremum operator, and \(\mathcal{F}\) denotes a class of functions used for the comparison. 

The empirical calculation of Maximum Mean Discrepancy (MMD) involves estimating the discrepancy between two distributions based on their samples. The formula for empirically calculating MMD is as follows:

\[
MMD(\mathcal{P}, \mathcal{Q}) = \frac{1}{{n(n-1)}} \sum_{{i \neq j}} k(x_i, x_j) + \frac{1}{{m(m-1)}} \sum_{{i \neq j}} k(y_i, y_j) - \frac{2}{{mn}} \sum_{{i, j}} k(x_i, y_j)
\]

where \(\mathcal{P}\) and \(\mathcal{Q}\) represent two distributions, \(x_i\) and \(y_i\) denote samples drawn from \(\mathcal{P}\) and \(\mathcal{Q}\) respectively, \(n\) and \(m\) are the respective sample sizes, and \(k\) is a kernel function.

The choice of the kernel function, \(k\), is crucial as it determines the sensitivity of MMD to different aspects of the distributions.

A commonly used kernel function is the Gaussian kernel, which measures the similarity between two samples based on their distance. The Gaussian kernel function is defined as:

\[
k(x, y) = \exp \left( -\frac{{\|x - y\|^2}}{{2\sigma^2}} \right)
\]

Here, \(x\) and \(y\) represent the samples from the distributions, and \(\sigma\) is a parameter controlling the width of the kernel.

By calculating the MMD between \(\mathcal{P}\) and \(\mathcal{Q}\), we can assess the dissimilarity between the distributions and detect covariate drift if the value of MMD is significant.

\subsubsection{Kolmogorov-Smirnov (KS) Statistic}
The KS statistic measures the maximum difference between the cumulative distribution functions of two probability distributions. It is commonly used for detecting changes in data distributions and can be applied to identify covariate drift in text data. By comparing the KS statistic between the distributions of the reference and current data, we can determine if there is a significant change in the data distribution, indicating covariate drift.

The KS statistic quantifies the maximum difference between the cumulative distribution functions (CDFs) of the two distributions being compared. Given two distributions, \(\mathcal{P}\) and \(\mathcal{Q}\), the KS statistic is computed as:

\[
KS = \max_{i} \left( \left| F_{\mathcal{P}}(x_i) - F_{\mathcal{Q}}(x_i) \right| \right)
\]

where \(F_{\mathcal{P}}(x_i)\) and \(F_{\mathcal{Q}}(x_i)\) represent the CDFs of \(\mathcal{P}\) and \(\mathcal{Q}\), respectively, and \(x_i\) denotes the \(i\)th data point.

In the context of multivariate data distributions, the KS statistic can be used by extending it to multiple dimensions. For each dimension, the KS statistic is calculated independently. Then, the maximum KS statistic across all dimensions is considered as the overall KS statistic for the multivariate data.

When comparing multiple dimensions simultaneously, it is important to account for multiple hypothesis testing to control the family-wise error rate. One commonly used correction method is the Bonferroni correction. The Bonferroni correction adjusts the significance threshold by dividing it by the number of dimensions being considered. This correction helps reduce the likelihood of false positive detections when performing multiple comparisons.

To apply the Bonferroni correction, suppose the desired significance level is \(\alpha\). If \(m\) dimensions are being compared, the adjusted significance level, denoted as \(\alpha_{adj}\), is given by:

\[
\alpha_{adj} = \frac{\alpha}{m}
\]

In hypothesis testing, the p-value is compared against the significance level to determine the statistical significance of the results. With the Bonferroni correction, the p-value threshold is adjusted as well. If the calculated p-value for a particular comparison is less than or equal to \(\alpha_{adj}\), it is considered statistically significant.

By applying the Bonferroni correction, the p-value threshold is made more stringent, reducing the chance of false positive detections. This correction is particularly useful in scenarios involving multiple comparisons, such as when comparing multiple dimensions in multivariate data, as it helps maintain the overall statistical validity of the analysis.

Here we report the p-value after multiplying it by m and we consider significance at 0.05 to follow standard convention.

By calculating the KS statistic and applying the Bonferroni correction, we can detect significant differences between distributions and identify covariate drift in multivariate data.

By combining these document embeddings, dimensionality reduction techniques, and drift detectors, we aim to evaluate the performance of different approaches in detecting covariate drift in text data.

\subsection{Dataset}
For our research, we utilize the AG-News dataset, specifically the AG-News Subset train dataset. The AG-News dataset is a widely used benchmark for text classification tasks, containing news articles from various categories. The subset we focus on consists of news articles from four categories: World, Sports, Business, and Sci/Tech.

The AG-News Subset is a labeled dataset that provides a balanced distribution of articles across the four categories. Each article in the dataset is represented by its title and description, capturing the essence of the news content. This dataset serves as a suitable choice for evaluating our methodology in detecting covariate drift in text data.

The AG-News Subset offers several advantages for our research. Firstly, it provides a diverse set of news articles covering different domains, enabling us to capture a wide range of textual variations and potential drift scenarios. Secondly, the balanced distribution of articles across categories ensures that our analysis is not biased towards any specific domain, allowing us to assess the performance of our methodology across different categories.

By employing the AG-News Subset, we aim to evaluate the effectiveness of our proposed approaches in detecting covariate drift and capturing distributional shifts in text data. The utilization of this dataset contributes to the robustness and generalizability of our findings.

\section{Experimental setup}

In this section, we describe the experimental setup used to evaluate our methodology for detecting covariate drift in text data. We constructed the training set and performed multiple experiments using the AG-News Subset train dataset.

\subsection{Datasets constructed}

We utilized the AG-News Subset dataset for our experiments.To ensure that our training set focuses on specific categories, we removed the sports category from the news. This allowed us to investigate the effectiveness of our methodology in detecting drift specifically related to sports news. The training set was constructed by randomly sampling 15,000 articles from the Ag news subset train dataset excluding the sports category. 

To evaluate the performance of our methodology, we created multiple test sets. Each test set consisted of 5,000 samples drawn randomly with replacement from the dataset, excluding the samples used in the training set. This sampling process was repeated five times to generate five distinct test sets for each experiment.

To evaluate the performance of our methodology in various drift scenarios, we modified the test datasets to include different percentages of sports news. Each test set still contained a total of 5,000 samples, but the proportion of sports news was varied. Specifically, we created test datasets with 0\%, 10\%, 25\%, 75\%, and 100\% of the samples being from the sports category. This modification refered as drift level below allowed us to assess the impact of different levels of sports news inclusion on the detection of covariate drift.

\subsection{Experimental Procedure}

For each experiment, we applied our methodology to detect covariate drift in the text data. We trained our models on the constructed training set and performed drift detection on each of the five drift levels separately.

To assess the statistical significance of the detected drift, we calculated the p-value for each test set. The p-value represents the probability of obtaining a test statistic as extreme as, or more extreme than, the observed value under the null hypothesis of no drift.

By repeating the experiment i.e each level of drift five times, we obtained a distribution of p-values. From this distribution, we computed the mean p-value as a measure of the average statistical significance across the multiple tests. Additionally, we calculated the standard deviation to quantify the variability of the results.

This experimental setup allowed us to evaluate the effectiveness of our methodology in detecting covariate drift in text data while providing robust statistical measures to support the validity of our findings.

For the code used to implement this experimental setup and perform the analysis, please refer to our GitHub repository at: \url{https://github.com/vinayaksodar/nlp_drift_paper_code.git}.

\section{Results and analysis}
The results of the experiments are presented in Tables 1-4, which show the p-values of the KS and MMD metrics for different models and drift levels. Significant p-values are highlighted in bold.

\begin{table}[h]
\centering
\begin{tabular}{c|cc|cc|c}
\hline
model & \multicolumn{2}{c|}{KS} & \multicolumn{2}{c|}{MMD} & drift level \\
 & mean & stddev & mean & stddev & \\
\hline
TFIDF-LSA & 0.05 & 0.05 & \textbf{0.00} & 0.00 & 0 \\
TFIDF-LSA & \textbf{0.00} & 0.00 & \textbf{0.00} & 0.00 & 0.10 \\
TFIDF-LSA & \textbf{0.00} & 0.00 & \textbf{0.00} & 0.00 & 0.25 \\
TFIDF-LSA & \textbf{0.00} & 0.00 & \textbf{0.00} & 0.00 & 0.50 \\
TFIDF-LSA & \textbf{0.00} & 0.00 & \textbf{0.00} & 0.00 & 0.75 \\
TFIDF-LSA & \textbf{0.00} & 0.00 & \textbf{0.00} & 0.00 & 1 \\
\hline
\end{tabular}
\caption{P-values of KS and MMD metrics for different percentages of samples using the tfidf model and dimentionality reduction using lsa. }
\label{tab:results}
\end{table}

\begin{table}[h]
\centering
\begin{tabular}{c|cc|cc|c}
\hline
model & \multicolumn{2}{c|}{KS} & \multicolumn{2}{c|}{MMD} & drift level \\
 & mean & stddev & mean & stddev & \\
\hline
doc2vec & 0.08 & 0.08 & 0.07 & 0.06 & 0 \\
doc2vec & \textbf{0.01} & 0.02 & \textbf{0.00} & 0.00 & 0.1 \\
doc2vec & \textbf{0.00} & 0.00 & \textbf{0.00} & 0.00 & 0.25 \\
doc2vec & \textbf{0.00} & 0.00 & \textbf{0.00} & 0.00 & 0.5 \\
doc2vec & \textbf{0.00} & 0.00 & \textbf{0.00} & 0.00 & 0.75 \\
doc2vec & \textbf{0.00} & 0.00 & \textbf{0.00} & 0.00 & 1 \\
\hline
\end{tabular}
\caption{P-values of KS and MMD metrics for different percentages of samples using the doc2vec model. }
\label{tab:results}
\end{table}

\begin{table}[h]
\centering
\begin{tabular}{c|cc|cc|c}
\hline
model & \multicolumn{2}{c|}{KS} & \multicolumn{2}{c|}{MMD} & drift level \\
 & mean & stddev & mean & stddev & \\
\hline
doc2vec-pca & 0.21 & 0.25 & 0.08 & 0.11 & 0 \\
doc2vec-pca & \textbf{0.05} & 0.05 & \textbf{0.01} & 0.01 & 0.1 \\
doc2vec-pca & \textbf{0.00} & 0.00 & \textbf{0.00} & 0.00 & 0.25 \\
doc2vec-pca & \textbf{0.00} & 0.00 & \textbf{0.00} & 0.00 & 0.5 \\
doc2vec-pca & \textbf{0.00} & 0.00 & \textbf{0.00} & 0.00 & 0.75 \\
doc2vec-pca & \textbf{0.00} & 0.00 & \textbf{0.00} & 0.00 & 1 \\
\hline
\end{tabular}
\caption{P-values of KS and MMD metrics for different drift levels using the doc2vec model and dimentionality reduction using PCA.}
\label{tab:results}
\end{table}

\begin{table}[h]
\centering
\begin{tabular}{c|cc|cc|c}
\hline
model & \multicolumn{2}{c|}{KS} & \multicolumn{2}{c|}{MMD} & drift level \\
 & mean & stddev & mean & stddev & \\
\hline
bert & 0.70 & 0.29 & 0.43 & 0.27 & 0 \\
bert & \textbf{0.00} & 0.00 & \textbf{0.00} & 0.00 & 0.1 \\
bert & \textbf{0.00} & 0.00 & \textbf{0.00} & 0.00 & 0.25 \\
bert & \textbf{0.00} & 0.00 & \textbf{0.00} & 0.00 & 0.5 \\
bert & \textbf{0.00} & 0.00 & \textbf{0.00} & 0.00 & 0.75 \\
bert & \textbf{0.00} & 0.00 & \textbf{0.00} & 0.00 & 1 \\
\hline
\end{tabular}
\caption{P-values of KS and MMD metrics for different percentages of samples using the bert model.}
\label{tab:results}
\end{table}

\begin{table}[h]
\centering
\begin{tabular}{c|cc|cc|c}
\hline
model & \multicolumn{2}{c|}{KS} & \multicolumn{2}{c|}{MMD} & drift level \\
 & mean & stddev & mean & stddev & \\
\hline
bert-pca & 0.61 & 0.46 & 0.44 & 0.26 & 0 \\
bert-pca & \textbf{0.00} & 0.00 & \textbf{0.00} & 0.00 & 0.1 \\
bert-pca & \textbf{0.00} & 0.00 & \textbf{0.00} & 0.00 & 0.25 \\
bert-pca & \textbf{0.00} & 0.00 & \textbf{0.00} & 0.00 & 0.5 \\
bert-pca & \textbf{0.00} & 0.00 & \textbf{0.00} & 0.00 & 0.75 \\
bert-pca & \textbf{0.00} & 0.00 & \textbf{0.00} & 0.00 & 1 \\
\hline
\end{tabular}
\caption{P-values of KS and MMD metrics for different drift levels using the bert model and dimentionality reduction using PCA. }
\label{tab:results}
\end{table}

 Among the drift detection metrics, the KS statistic performs surprisingly well by detecting drift at all drift levels in all the experiments. It even provides higher p-values than MMD when there is no supposed drift. The KS statistic is also computationally more efficient as calculating a p-value for it doesn't require a permutation test, unlike MMD. However, MMD is also able to detect drift at all levels. It performs worse when there is no drift and even detects drift at the 0 drift level in the TFIDF-LSA experiment.

Among the models used, the bert model performed the best across metrics, regardless of whether dimensionality reduction was used. The TFIDF model performed the worst, while the doc2vec model was somewhere in the middle. Interestingly, dimensionality reduction doesn't seem to impact the results significantly in both the doc2vec and bert models.

\section{Conclusion}
The results obtained from these experiments provide valuable insights into the performance of different models and drift detection metrics. The KS statistic proves to be a reliable metric, consistently detecting drift across all experiments. MMD, while effective in detecting drift, has limitations when there is no actual drift present. The bert model stands out as the top performer, indicating its robustness in capturing and adapting to drift in text data. On the other hand, the TFIDF model demonstrates weaker performance, suggesting the need for more sophisticated approaches in detecting drift. The doc2vec model performs reasonably well, positioning it between the TFIDF and bert models. Since dimentionality reduction seems doen't seem to impact the results, it may be used when it is more computationally efficient to do so

These findings contribute to the understanding of drift detection in text data. Further analysis and experimentation can be conducted to explore additional models, dimensionality reduction techniques, and drift detection methods to improve the accuracy and efficiency of drift detection in various text analysis tasks.


\end{document}